\documentclass{article}

\PassOptionsToPackage{numbers,sort&compress}{natbib}

\usepackage[preprint]{neurips_2020}

\usepackage[utf8]{inputenc} 
\usepackage[T1]{fontenc}    
\usepackage{hyperref}       
\usepackage{url}            
\usepackage{booktabs}       
\usepackage{amsfonts}       
\usepackage{nicefrac}       
\usepackage{microtype}      
\usepackage{amsmath}

\usepackage{bbm}
\DeclareMathOperator*{\argmin}{argmin}

\newcommand{\std}[1]{{\scriptsize{$\pm$#1}}}

\renewcommand{\paragraph}[1]{\textbf{#1}\hspace{1em}}

\usepackage{graphicx}
\graphicspath{{figures/}}
\usepackage{subcaption}
\usepackage{multirow}

\usepackage{chngcntr}
\usepackage{float}

\title{Subject-Aware Contrastive Learning for Biosignals}

\author{%
  Joseph Y.~Cheng, Hanlin Goh, Kaan Dogrusoz, Oncel Tuzel, Erdrin Azemi \\
  Apple \\
  \texttt{\{\href{mailto:jycheng@apple.com}{jycheng},\,\href{mailto:hanlin@apple.com}{hanlin},\,\href{mailto:kdogrusoz@apple.com}{kdogrusoz},\,\href{mailto:otuzel@apple.com}{otuzel},\,\href{mailto:erdrin@apple.com}{erdrin}\}@apple.com}
}

\begin{document}

\maketitle


\begin{abstract}
Datasets for biosignals, such as electroencephalogram (EEG) and electrocardiogram (ECG), often have noisy labels and have limited number of subjects ($<$100). To handle these challenges, we propose a self-supervised approach based on contrastive learning to model biosignals with a reduced reliance on labeled data and with fewer subjects. In this regime of limited labels and subjects, intersubject variability negatively impacts model performance. Thus, we introduce subject-aware learning through (1)~a \textit{subject-specific} contrastive loss, and (2)~an adversarial training to promote \textit{subject-invariance} during the self-supervised learning. We also develop a number of time-series data augmentation techniques to be used with the contrastive loss for biosignals. Our method is evaluated on publicly available datasets of two different biosignals with different tasks: EEG decoding and ECG anomaly detection. The embeddings learned using self-supervision yield competitive classification results compared to entirely supervised methods. We show that subject-invariance improves representation quality for these tasks, and observe that subject-specific loss increases performance when fine-tuning with supervised labels.

\end{abstract}

\section{Introduction}

Biosignal classification can lead to better diagnosis and understanding of our
bodies and well-being. For example, medical experts can monitor health
conditions, such as epilepsy \cite{acharya_automated_2013} or depression
\cite{de_aguiar_neto_depression_2019}, using brain electroencephalography (EEG)
data. In addition, electrocardiograms (ECG) give insight to cardiac health and
can also indicate stress \cite{healey_detecting_2005}.

Time-series biosignals can be non-invasively and continuously measured. However,
labeling these high-dimensional signals is a labor-intensive and time-consuming
process, and assigning labels may introduce unintended biases. We propose to use
self-supervised learning methods to extract meaningful representations from
these high-dimensional signals without the need of labels. In this work, we make the following contributions:

\paragraph{Apply self-supervised learning to biosignals.} Speech and vision
data are processed by the human senses of hearing and sight. Whereas, biosignals
can be the result of processing this information along with other complex
biological mechanisms. Information may be obscured or lost when measuring these
processes with the resulting time-series signals. Thus, it is unclear whether
the same techniques used to learn representations for language and vision
domains are effective for biosignals. In this work, we demonstrate the
effectiveness of contrastive loss for self-supervised learning for biosignals.
This approach requires the development and assessment of augmentation techniques
and the consideration of intersubject variability, the signal variations from
subject to subject.

\paragraph{Develop data augmentation techniques for biosignals.} Data
transformation algorithms help increase the information content in the learned
embeddings for desired downstream tasks. In this work, we develop
domain-inspired augmentation techniques. For example, the power in certain EEG
frequency bands has been shown to be highly correlated with different brain
activities. Thus, we use frequency-based perturbations to augment the signal. We
find that temporal specific transformations (cutout and delay) are the most
effective transformations for representation learning followed by signal mixing,
sensor perturbations (dropout and cutout), and bandstop filtering.

\paragraph{Integrate subject awareness into the self-supervised learning
framework.} Inter-subject variability poses a challenge when performing
data-driven learning with a small number of subjects. Subject-specific features
are an integral part of biosignals, and the knowledge that the biosignals are
from different subjects is a ``free'' label. We investigate two solutions to
integrate this feature into the self-supervised learning framework: (1) using
subject-specific distributions to compute the contrastive loss, and (2)
promoting subject invariance through adversarial training. Experimental results
show that promoting subject invariance increases classification performance when
training with a small number of subjects. Both approaches yield weight
initializations that are effective in fine-tuning with supervised labels.
\section{Related Work}

\paragraph{Representation learning for temporal signals.} Temporal signals
introduce an additional dimension to exploit for self-supervised learning.
Previous work used temporal dynamics such as temporal ordering
\cite{misra_shuffle_2016,wei_learning_2018} or temporal prediction
\cite{oord_representation_2019}. Designing and including multiple pretext tasks
\cite{doersch_multi-task_2017} has been shown to be effective in representation
learning for both EEG \cite{banville_self-supervised_2019} and ECG signals
\cite{sarkar_self-supervised_2020}.

Alternatively, representation learning can be performed through clustering with
contrastive learning
\cite{wu_unsupervised_2018,sermanet_time-contrastive_2018,he_momentum_2019,chen_simple_2020}.
In this framework, a single instance is clustered with different perturbations
of itself while pushing all other instances away in the embedding space. The
non-stationary property of biosignals can be exploited where further time
instances of the same data stream can be considered as negative training
examples \cite{hyvarinen_unsupervised_2016}. Here, we take the concepts proposed
by Hyv\"arinen and Morioka \cite{hyvarinen_unsupervised_2016} that was
demonstrated for magnetoencephalography (MEG), and we apply the contrastive
learning framework \cite{chen_simple_2020} for time-series biosignals. This
application requires careful design of appropriate data transformations and
consideration of intersubject variability.

\paragraph{Incorporating subject-based information.} Intersubject variability
can be interpreted as a different domain for each subject. The domain shift from
subject to subject can be modeled and corrected with adversarial training
\cite{tzeng_adversarial_2017}, as shown for neural signals recorded with
multi-electrode arrays \cite{farshchian_adversarial_2019}. The model can also be
promoted to learn domain-invariant features
\cite{tzeng_simultaneous_2015,xie_controllable_2017} as demonstrated for
regularizing a classifier for EEG processing \cite{ozdenizci_learning_2020}. In
our work, we introduce subject-invariance into the contrastive learning
framework and investigate the impact using the learned representations in
downstream tasks.
\section{Methods}

Self-supervised learning is performed using contrastive learning with data
transformation techniques
\cite{wu_unsupervised_2018,he_momentum_2019,chen_simple_2020}. First,
contrastive learning framework is described with the proposed addition of
subject awareness to address intersubject variability. Next, specific
transformation techniques for training are designed and introduced to more
effectively learn biosignal representations. Lastly, the proposed approach is
applied to downstream tasks as described in the next section.

\subsection{Model and Contrastive Loss Function}

\begin{figure}
  \centering
  \includegraphics[width=\linewidth]{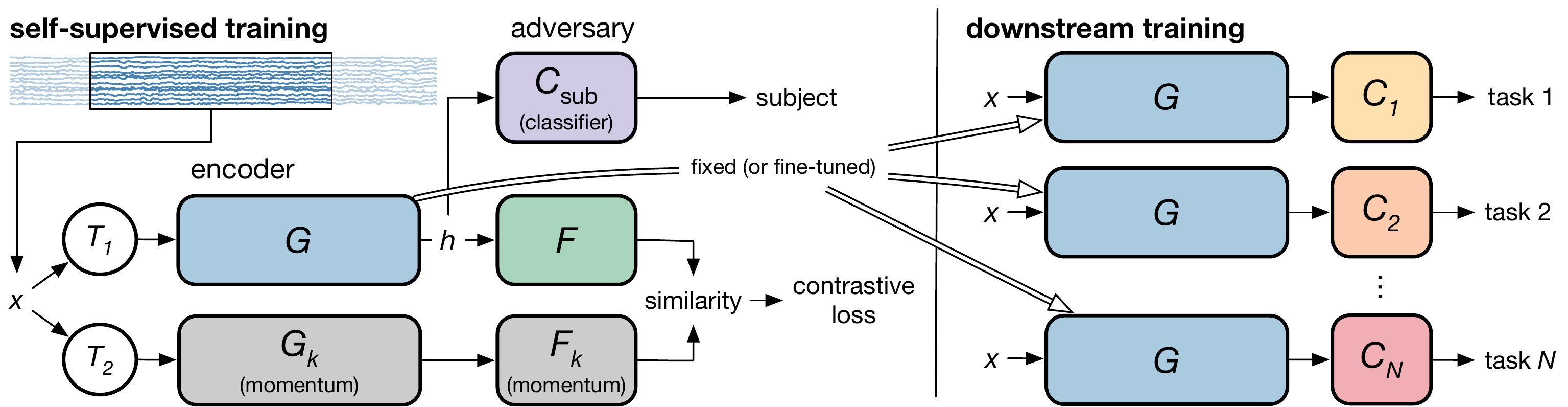}
  \caption{Method overview. An encoder $G$ is trained using self-supervised
  training (left) with an adversarial subject identifier to minimize
  subject-specific content. Alternatively, subject-awareness can be introduced
  through subject-specific contrastive loss. The resulting encoder can then be
  applied for different downstream tasks by attaching a classification model
  (right).}
  \label{fig:method}
\end{figure}

The method is summarized in Fig.~\ref{fig:method}. Encoder $G(T_1(x_i))$ with
parameters $\theta_G$ encodes data sample $x_i$ with transformation $T_1$ to
latent representation $h_i$. During self-supervised training, $h_i$ is passed
through model $F(h_i)$ with parameters $\theta_F$ to yield $q_i$: $q_i =
F\left(G\left(T_1\left(x_i\right)\right)\right)$. Copies of $G$ and $F$ as $G_k$
and $F_k$ are used to generate $k_i$ from $x_i$ with transformation $T_2$: $k_i
= F_k(G_k(T_2(x_i))$. The models $G_k$ and $F_k$ are updated with momentum
\cite{he_momentum_2019}. Both $q_i$ and $k_i$ are normalized to unit L2-norm.

The models are trained to maximize the mutual information between $T_1(x_i)$ and
$T_2(x_i)$ for any $T_1$ and $T_2$. The mutual information maximization is
estimated with InfoNCE
\cite{gutmann_noise-contrastive_2010,oord_representation_2019}:
\begin{equation}
  \ell_{i} = -\log{\frac{\exp{\left(q_i^T k_i / \tau\right)}}{\sum_{j=0}^N\exp{\left(q_i^Tk_j/\tau\right)}}},
  \label{eq:infonce}
\end{equation}
where the inner product $q_i^Tk_i$ is used as a similarity metric and $\tau$ is
a learnable temperature parameter. In Eq.~\ref{eq:infonce}, $q_i^T k_i$ is
contrasted against the inner product of $q_i$ and $N-1$ negative examples. The
momentum update of $G_k$ and $F_k$ enables the use of negative examples from
previous batches to increase the number of negative examples
\cite{he_momentum_2019}.

This loss focuses on features that differentiate each time segment from other
time segments. With the variability in how the sensors are setup and how the
signals behave for each subject, the learned embeddings will be dominated by
subject-specific characteristics. To promote the extraction of subject-invariant
features, we apply adversarial training \cite{goodfellow_generative_2014,
xie_controllable_2017}. A classifier $C_\textrm{sub}$ is trained to predict the
identity of the subject of each example based on latent vector $h_i$:
$C_\textrm{sub}(h_i) = C_\textrm{sub}(G(x_i))$. The $j$-th element,
$C^j_\textrm{sub}(h_i)$, corresponds to the probability of $h_i$ being from
subject $j$. This model is trained with a fixed encoder $G$ using cross entropy
loss:
\begin{equation}
  \ell_{\textrm{sub},i} = -\sum_{j=0}^{N_\textrm{sub}-1} \mathbbm{1}_{[j = s_i]} \log C^j_\textrm{sub}(G(x_i)),
\end{equation}
where $N_\textrm{sub}$ is the number of subjects, $s_i$ is the subject number of
example $i$, and $\mathbbm{1}_{[j = s_i]}$ is an indicator function with a value
of 1 when $j = s_i$. The encoder $G$ is encouraged to confuse this subject
classifier by using a fixed $C_\textrm{sub}$ to regularize the training of $G$
and $F$ with the regularization term of:
\begin{equation}
  r_{\textrm{sub}, i} = -\sum_{j=0}^{N_\textrm{sub}-1} \mathbbm{1}_{[j = s_i]} \log \left(1 - C^j_\textrm{sub}(G(x_i)) \right).
\end{equation}
The final loss functions become:
\begin{equation}
  \argmin_{F, G} \mathbbm{E}_i\left[ \ell_i + \lambda r_{\textrm{sub}, i} \right], \textrm{and } \argmin_{C} \mathbbm{E}_i\left[ \ell_{\textrm{sub},i} \right],
  \label{eq:loss:adv}
\end{equation}
where $\lambda$ is a tunable hyperparameter. Unless specified, $\lambda$ was set
to 1 in the experiments. We refer to this approach as \textit{subject-invariant}
self-supervised learning since the embedding is regularized to minimize
subject-based information. For multi-session measurements, the differences in
sensor setup for the same subject will also be a factor but can be treated as a
separate subject with this formulation.

Subject information can also be incorporated in the negative sampling procedure
when computing the contrastive loss. The noise estimation can be based only on
the data distribution of a single subject. This procedure focuses the loss on
differences in time for a single subject, and does not consider differences
between subjects. We refer to this approach as \textit{subject-specific}
self-supervised learning. We refer to self-supervised learning (SSL) without
subject-awareness as \textit{base SSL}.

\subsection{Data Transformations for Biosignals}

\begin{figure}
  \centering
  \includegraphics[width=\linewidth]{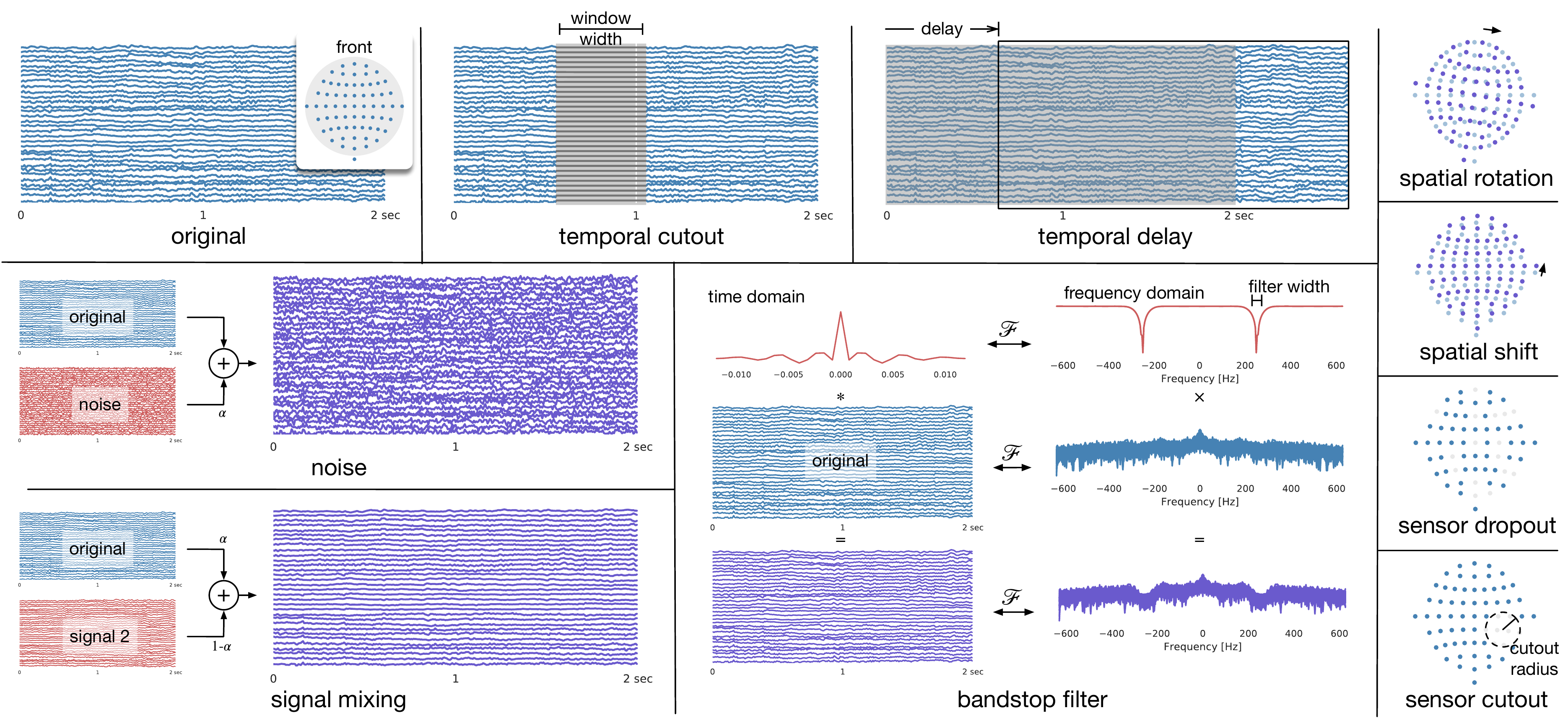}
  \caption{Data transformations for augmentations demonstrated for EEG signals.
  The original data input is shown in the upper left corner. Multi-channel
  biosignals can be further augmented with spatial perturbations (right column).
  All augmentations were used for our experiments with the exception of spatial
  rotation/shift and sensor cutout for ECG.}
  \label{fig:augmentation}
\end{figure}

Two random transformations are applied to each training example $x_i$: $y_{i,1}
= T_1(x_i)$ and $y_{i,2} = T_2(x_i)$. We investigate a number of transformations
in the context of biosignals (Fig.~\ref{fig:augmentation}):
\begin{itemize}
  \setlength{\itemsep}{0pt}
  \item \textit{Temporal cutout:} a random contiguous section of the time-series
  signal (cutout window) is replaced with zeros \cite{devries_improved_2017};
  \item \textit{Temporal delays:} the time-series data is randomly delayed in time;
  \item \textit{Noise:} independent and identically distributed Gaussian noise is added to the signal;
  \item \textit{Bandstop filtering:} the signal content at a randomly selected
  frequency band is filtered out using a bandstop filter; and
  \item \textit{Signal mixing:} another time instance or subject data is added to the signal to simulate correlated noise.
\end{itemize}

These signals are collected from physical sensors that are dispersed in space.
For EEG signals, multiple electrodes are placed around the surface of the head.
These spatial locations can be used to help augment the data with
domain-specific transformations:
\begin{itemize}
  \setlength{\itemsep}{0pt}
  \item \textit{Spatial rotation:} the data is rotated in space \cite{krell_rotational_2017};
  \item \textit{Spatial shift:} the data is shifted in space;
  \item \textit{Sensor dropout:} a random subset of sensors is replaced with zeros; and
  \item \textit{Sensor cutout:} sensors in a small region of space are replaced
  with zeros.
\end{itemize}

For random bandstop filters, we designed a fixed low-pass filter using a hamming
window with 31 coefficients. This filter was first converted to a bandpass
filter by modulating the filter using a cosine function with a random phase
selected with uniform probability. The signal was bandpassed using this filter
and subtracted from the original signal. Radial basis function interpolation was
used to perform the spatial rotation and spatial shift transformations.

\section{Experiments and Results}

All experiments were performed using PyTorch (v1.4.0) \cite{NIPS2019_9015}.
One-dimensional ResNet models \cite{he_identity_2016} with ELU activation and
batch normalization \cite{ioffe_batch_2015} were used for encoders $G$ with
slightly different parameters for each application (see Appendix for details).
Model $F$ consisted of a 4-layer fully-connected network with 128 dimensions at
each stage and 64 dimensions at the output. Unless specified, Adam optimizer
\cite{kingma_adam_2017} was used with a learning rate of 1e-4. Self-supervised
learning with momentum was applied with a $k_i$ history of 24k elements and an
update momentum of 0.999. On an NVIDIA Tesla V100 card with 32GB, this training
took 47.8--54.6 hours for the EEG dataset and 33.7--39.2 hours for the ECG
dataset. Linear classification using logistic regression with weight decay of
0.01 was performed to evaluate the quality of the learned embeddings. Results
were reported as mean $\pm$ standard deviation across 10 trials, each performed
with a different random seed.

\subsection{EEG: PhysioNet Motor Imagery Dataset}

Electrical neural activity can be non-invasively recorded using electrodes
placed on the scalp with EEG. Being able to derive meaningful representations
from these signals will enable further understanding of the brain. However,
these signals are difficult to interpret and label. Therefore, we applied our
approach to the PhysioNet motor movement/imagery dataset
\cite{goldberger_physiobank_2000,schalk_bci2000_2004,physionet_eeg_url}. Data
were recorded from 109 volunteers where each subject was asked to perform
imagined motor tasks: closing the right fist, closing the left fist, closing
both fists, and moving both feet. Each task lasted for $\sim$4 sec and was also
performed with actual movement. Following previous work \cite{kim_motor_2016},
we excluded data from 3 volunteers due to inconsistent timings. During the
experiment, 64-channel EEG data were recorded at 160~Hz using the BCI2000 system
\cite{schalk_bci2000_2004}. We re-referenced the raw data using the channel
average, and normalized the data by the mean and standard deviation of the
training dataset.

\paragraph{Experimental Setup.} Encoder $G$ (288k parameters, model details in
Appendix) was trained using self-supervised learning with data from 90 subjects.
For each recording (both imagined and action trials), a time window was randomly
selected as the input to $G$. A 256-dimensional embedding vector was produced by
the encoder for every 320 samples (2 sec). Self-supervised training was
performed with a batch size of 400 and 270k steps (or 54k steps if only one
augmentation or none was applied).

A logistic-regression linear classifier was trained on top of the frozen $G$
model using the class labels and data from the same 90 subjects for the imagined
trials only. Each example was a 2-sec window of data selected 0.25 sec after the
start of each trial to account for response time after the cue. The resulting
classifier and encoder $G$ was then evaluated on 16 held-out subjects. This
experimental setup is referred as \textit{intersubject testing}. Two downstream
task setups were performed: (1)~2-class problem of determining imagined right
fist or left fist movements based on the EEG signal, and (2)~4-class problem of
determining imagined right fist, left fist, both fists, or both feet movements.
Classifiers were trained using a learning rate of 1e-3 and batch size of 256 for
2k epochs.

\paragraph{Impact of Transformation.}
To understand the effectiveness of each transformation (e.g.~temporal cutout,
random temporal delay) and the associated parameter (e.g.~temporal cutout
window, maximum temporal delay), a single transformation type was applied for
$T_2$ during the self-supervised training, and the identity transform was used
for $T_1$. Afterwards, the learned encoder $G$ was evaluated by training a
linear classifier on top of the frozen network for the 4-class task.

Temporal cutout was the most effective transformation followed by temporal delay
and signal mixing (Table~\ref{table:eeg:aug:abbrev}). The effect of temporal
transformations was the promotion of temporal consistency where neighboring time
points should be close in the embedding space and more distant time points
should be farther. This finding was in agreement with previous work that
considered the non-stationary property of biosignals and exploited this property
with time contrastive learning \cite{hyvarinen_unsupervised_2016}.
Less effective were spatial perturbations with negligible improvement
($\leq$0.1\%) in accuracy (not shown) --- likely the result of the limited
spatial resolution of the EEG modality.


\begin{table}[h] \caption{Impact of transformation type on representation
  learning evaluated with a linear classifier trained on top of the learned
  encoder for the EEG dataset. Accuracies are computed for the 4-class problem
  and are shown as differences compared to the baseline of training with no
  augmentation (36.7\%). Best parameter for each transformation is
  shown. See Appendix for a more detailed analysis.}
  \label{table:eeg:aug:abbrev}
  \centering
  \small
  \begin{tabular}{lcccccccc}
    \toprule
    & Max time & Temporal & Noise & Bandstop & Mixing & Sensor      & Sensor cutout \\
    & delay    & cutout   & scale & width    & scale  & dropout $p$ & radius$^a$ \\
    \hline
    Parameter
    & 40 samp & 200 samp & 6.0 & 64 Hz & 0.9 & 0.2 & 0.25 \\
    Accuracy [\%]
    & +6.1\std{1.2} & +10.9\std{2.0} & +3.8\std{0.9} & +3.7\std{1.5}
    & +5.3\std{1.4} & +3.8\std{1.5} & +4.1\std{1.4}\\
    \bottomrule
  \end{tabular}\\
  \footnotesize{$^a$Spatial units are normalized such that the entire coverage
  in all directions is between 0 and 1.}
\end{table}

\paragraph{Impact of Subject-Aware Training.}
The impact of subject-aware training was evaluated by performing the
self-supervised training with different configurations. Randomly-initialized
encoder was used as a baseline for comparison. Applying SSL with no augmentation
was no better than using this random encoder. For intersubject testing, the
different variants of SSL performed comparably (Table~\ref{table:eeg:losses}).
The training set was sufficiently large (90 subjects) to generalize to unseen
subjects.


\begin{table}
  \caption{Ablation evaluation of learned EEG representation using a linear
  classifier with a frozen encoder. Both subject-specific and subject-invariant
  training decreased subject features (lower subject identification accuracies)
  and improved classification accuracies.}
  \label{table:eeg:losses}
  \centering
  \small
  \begin{tabular}{lccccc}
    \toprule
    & \multicolumn{2}{c}{Intersubject} & \multicolumn{3}{c}{Intrasubject} \\
    \cmidrule(lr){2-3}\cmidrule(lr){4-6}
    & 2 class$^a$ & 4 class$^b$ & 2 class$^a$ & 4 class$^b$ & Sub ID$^c$\\
    \hline
    No augmentation
    & 67.7\std{1.0} & 36.7\std{1.4}
    & 62.4\std{2.8} & 31.9\std{2.2}
    & 48.8\std{1.9} \\

    Base SSL
    & 78.8\std{1.0} & 49.4\std{1.0}
    & 77.6\std{2.0} & 46.6\std{2.9}
    & \textbf{88.6\std{2.1}} \\


    Subject-specific
    & \textbf{79.1\std{1.0}} & \textbf{49.9\std{0.4}}
    & 77.2\std{2.8} & 46.4\std{1.5}
    & 68.4\std{2.4} \\

    Subject-invariant
    & 79.0\std{0.5} & 49.8\std{0.7}
    & \textbf{79.4\std{2.1}} & \textbf{50.3\std{2.2}}
    & 73.0\std{1.8} \\


    \hline
    Random encoder
    & 68.0\std{1.7} & 36.3\std{1.6}
    & 66.3\std{2.0} & 32.9\std{1.7}
    & 55.9\std{2.7} \\
    \bottomrule
  \end{tabular}\\
  \footnotesize{$^a$2 class: right fist and left fist; $^b$4 class: right
  fist, left fist, both fists, and both feet; $^c$identifying 16 subjects.}
\end{table}

We also performed \textit{intrasubject testing} where nonoverlapping portions of
the data from the same set of 16 subjects were used for training (75\% of data)
and testing (25\%). The subjects that were not used for the self-supervised
training were used for training and testing the linear classifier. This setup
simulated the scenario where labels are available for new subjects from a
calibration process. In this scenario, performance increased for the
subject-invariant encoder. The greatest improvement was observed for 4 classes
with 50.3\% accuracy. We believe that this increase was due to minimizing the
impact of subject variability through subject-invariant training.

\begin{figure}
  \begin{minipage}[c]{0.49\textwidth}
    \includegraphics[width=\linewidth]{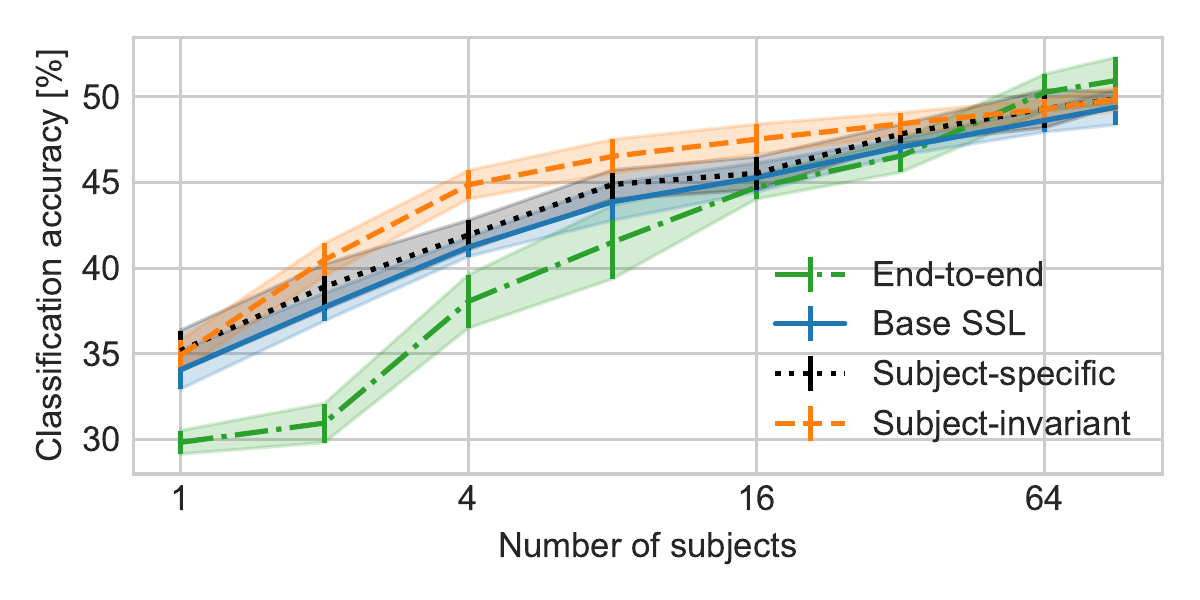}
  \end{minipage}
  \hfill
  \begin{minipage}[c]{0.5\textwidth}
    \caption{Intersubject testing for 4 classes for the EEG dataset. A linear
    classifier was trained on top of the frozen encoder using $N$ subjects
    ($x$-axis) and was tested on 16 unseen subjects. Supervised end-to-end
    learning was used as a baseline for reference. Fewer subjects were needed
    for self-supervised models: using labels from 4 subjects for
    subject-invariant SSL resulted in similar performance to end-to-end learning
    with 16 subjects.}
    \label{fig:eeg:labels}
  \end{minipage}
\end{figure}

\paragraph{Impact of Fewer Labels.}
We also investigated whether this self-supervised learning approach provided the
ability to learn tasks with fewer labels (Fig.~\ref{fig:eeg:labels}). With fewer
subjects to train the classifier, the subject-invariant SSL produced an encoder
that was less impacted by subject variability, as seen by the performance over
the base SSL. With enough subjects used to train the classifier, subject
variability became less problematic; the training examples sufficiently covered
different types of subjects to generalize to new subjects. For larger number of
subjects ($\sim$64), the base SSL performed comparably to the subject-invariant
SSL. For intrasubject testing, the subject-invariant SSL consistently produced a
better performing encoder compared to all other variants of SSL and supervised
end-to-end learning for these 16 subjects regardless of the percentage of labels
used (figure in Appendix).

\paragraph{Fine-Tuning with Supervised Labels.}
The models produced from self-supervised learning were evaluated as an
initialization for fine-tuning with supervised labels
(Table~\ref{table:eeg:finetune}). A fully connected layer was attached to the
last encoder layer, and this layer along with the entire encoder was fine-tuned
(learning rate of 1e-5, batch size of 256, and 200 epochs). For intersubject
classification, we achieved 81.6\% for 2 classes, and 53.9\% for 4 classes.
The increased accuracies with using self-supervised-trained models may be
attributed to using more data (both action and imagined trials) for training the
initial encoder.  Reducing subject information in the encoder (lower
classification accuracies for subject identification) provided a better
initialization for the EEG motor imagery task.


\begin{table}
  \centering
  \caption{EEG classification accuracy [\%]. General representations were
  learned through self-supervised training, and fine-tuning from these models
  (last three rows) improved accuracies compared to end-to-end (ours) training
  from random initialization.}
    \label{table:eeg:finetune}
  \small
  \begin{tabular}{lcccc}
    \toprule
    & \multicolumn{2}{c}{Intersubject} & \multicolumn{2}{c}{Intrasubject} \\
    \cmidrule(lr){2-3}\cmidrule(lr){4-5}
    & 2 class$^a$ & 4 class$^b$ & 2 class$^a$ & 4 class$^b$ \\
    \hline
    Kim et al., Random forest \cite{kim_motor_2016} & 80.1 & - & - & - \\
    Dose et al., CNN \cite{dose_deep_2018} & 80.1 & - & - & - \\

    End-to-end (ours)
    & 81.0\std{0.9} & 50.6\std{1.0}
    & 76.3\std{2.4} & 44.0\std{1.6}\\
    \hline
    \textit{Fine-tuned from:}\\

    \hspace{10pt}Base SSL
    & 81.1\std{0.6} & 52.6\std{0.7}
    & 78.6\std{2.5} & \textbf{50.8\std{1.4}} \\

    \hspace{10pt}Subject-specific
    & \textbf{81.6\std{0.8}} & \textbf{53.9\std{0.4}}
    & 79.3\std{2.0} & 50.5\std{1.6}\\

    \hspace{10pt}Subject-invariant
    & 81.2\std{0.9} & 52.8\std{0.8}
    & \textbf{79.6\std{2.3}} & 49.8\std{1.5} \\
    \bottomrule
  \end{tabular}\\
  \footnotesize{$^a$2 class: right fist and left fist; $^b$4 class: right fist,
  left fist, both fists, and both feet.}
\end{table}

\subsection{ECG: MIT-BIH Arrhythmia Database}

We also evaluated our approach on ECG signals. These signals assist in the
detection and characterization of cardiac anomalies on the beat-by-beat level
and on the rhythm level. ECG datasets have different challenges for data-driven
learning: these datasets are highly imbalanced, and features for detecting
anomalies may be tied closely to the subject. We investigated the impact of our
methods in these situations. We used the MIT-BIH Arrhythmia Database
\cite{goldberger_physiobank_2000, moody_impact_2001,physionet_ecg_url} that is
commonly used to benchmark ECG beat and rhythm classification algorithms. This
dataset contained 48 ambulatory ECG recordings from 47 different subjects. The
30-min recordings from two sensor leads were digitized to 360 Hz with a bandpass
filtering of 0.1--100 Hz. Signals were annotated by expert cardiologists to
denote the type of cardiac beat and cardiac rhythms.

\paragraph{Experimental Setup.}
The setup described by de Chazal et al.~\cite{de_chazal_automatic_2004} was
used. Recordings were divided into a training set and testing set where the
different types of beats and rhythms were evenly distributed: 22 recordings in
training set, 22 recordings from different subjects in the testing set, and 4
excluded recordings due to paced beats. These 4 excluded recordings were
included in the self-supervised learning. Cardiac beats were categorized into 5
classes: normal beat (training / testing samples of 45.8k / 44.2k),
supraventricular ectopic beat (SVEB, 0.9k / 1.8k), ventricular ectopic beat
(VEB, 3.8k / 3.2k), fusion beat (414 / 388), and unknown beat (8 / 7). The
dataset was highly imbalanced, and thus, we followed the common practice of
training a 5-class classifier and evaluating its performance in terms of
classifying SVEB and VEB. To evaluate different setups, balanced accuracies were
computed without the unknown beat due to too few examples for training and
testing. The dataset was also labeled with rhythm annotations
\cite{dash_automatic_2009,bruun_automatic_2017,wu_deep_2019}: normal sinus
rhythm (training / testing samples of 3.3k / 2.8k), atrial fibrillation (195 /
541), and other (256 / 362).

For an input window of 704 samples (1.96 sec), a 256-dimensional vector was
produced from the encoder $G$ (985k parameters, model details in Appendix). The
256-dimensional vector was used directly to train a linear classifier for beat
classification. For rhythm classification, 5 segments (9.78 sec) of data
produced 5 vectors that were average pooled into a single vector before applying
a linear classifier. Each window of ECG data $x$ was centered by the mode $m(x)$
and normalized by $\mathbb{E}[(x - m(x))^2]^{0.5}$. A batch size of
1000 and 260k steps were used for self-supervised training.

\paragraph{Impact of Subject-Aware Training.} Different self-supervised learning
setups were used to assess the impact of subject-aware training. Because subject
characteristics were closely tied to the beat and rhythm classes, regularization
parameter $\lambda$ for subject-invariant training was varied from 0.001 to 1.0.
To evaluate the quality of the learned embeddings, a linear classifier was
trained on top of the frozen encoder using cross entropy (weight decay of 0.01,
learning rate of 1e-3, batch size of 256, and 1k epochs). For rhythm
classification, training data had 90\% overlap for augmentation; no overlap was
used for testing. Examples were randomly sampled with replacement to account for
class imbalance.

Similar to EEG, subject-invariant contrastive learning produced the best
performing representation for ECG beat and rhythm classification
(Table~\ref{table:ecg:inter}). In this case, the subject-invariant
regularization $\lambda$ was lowered to 0.1 and 0.01 to maintain sufficient
subject information for beat and rhythm classifications. A lower regularization
was required due to the uneven distribution of labels between each subject.


\begin{table}
  \caption{Linear classifiers to evaluate the information in the frozen
  encoders. Subject-invariance with minimal subject-based features performed
  best for classifying beats ($\lambda=$~0.01) and rhythms (0.001) in terms of overall balanced accuracies.}
  \label{table:ecg:inter}
  \centering
  \small
  \begin{tabular}{lcccccc}
    \toprule
    & \multicolumn{3}{c}{Beat}
    & \multicolumn{2}{c}{Rhythm}
    & Subject \\
    \cmidrule(lr){2-4}
    \cmidrule(lr){5-6}
    \cmidrule(lr){7-7}
    & Overall
    & \multicolumn{1}{c}{SVEB} & \multicolumn{1}{c}{VEB}
    & Overall & AFib
    & ID \\
    & Bal Acc & F1 & F1 & Bal Acc & F1 & Acc \\
    \hline

    No augmentation
    & 47.9\std{3.8} & 6.9\std{2.0} & 34.1\std{3.4}
    & 38.0\std{1.5} & 41.5\std{4.3}
    & 67.5\std{1.8} \\

    Base SSL
    & 66.3\std{6.2} & \textbf{26.7\std{8.3}} & 56.1\std{8.3}
    & 49.4\std{3.4} & 39.7\std{6.1}
    & \textbf{84.7\std{0.6}} \\

    Subject-specific
    & 65.1\std{6.1} & 20.9\std{7.7} & 57.4\std{5.9}
    & 46.0\std{3.7} & 39.4\std{5.3}
    & 81.6\std{0.8} \\

    Subject-invariant \\
    \hspace{5pt}$\lambda=$~1.0
    & 64.3\std{7.9} & 18.7\std{8.5} & 59.9\std{9.2}
    & 49.7\std{3.3} & 42.9\std{8.4}
    & 76.1\std{2.2} \\

    \hspace{5pt}$\lambda=$~0.1
    & 68.6\std{6.0} & 26.5\std{8.1}  & 62.9\std{10.9}
    & 51.0\std{4.4} & \textbf{44.8\std{7.9}}
    & 79.0\std{0.9}\\

    \hspace{5pt}$\lambda=$~0.01
    & \textbf{69.8\std{5.5}} & 23.9\std{8.4} & \textbf{64.1\std{8.5}}
    & 52.3\std{5.9} & 44.6\std{5.8}
    & 82.9\std{0.9} \\

    \hspace{5pt}$\lambda=$~0.001
    & 63.5\std{4.4} & 19.7\std{5.3} & 62.0\std{8.0}
    & \textbf{52.4\std{4.3}} & 39.8\std{6.1}
    & 83.7\std{0.9}\\

    \hline
    Random encoder
    & 54.9\std{3.8} & 5.6\std{0.9} & 61.5\std{11.3}
    & 37.0\std{2.0} & 40.7\std{2.8}
    & 41.4\std{3.5} \\
    \bottomrule
  \end{tabular}\\
  \footnotesize{Abbreviations: SVEB = supraventricular ectopic beat; VEB =
  ventricular ectopic beat; AFib = atrial fibrillation; ID = subject
  identification; Bal Acc = balanced accuracy [\%]; Acc = accuracy [\%]; F1 = F1
  score; ID = subject identification (trained with 5 min from all subjects and
  tested on unseen data of 22 subjects not used for SSL).}
\end{table}

\paragraph{Impact of Fewer Labels.} Using the self-supervised model, fewer
labels may potentially be needed. The learned encoder was frozen, and a linear
classifier was trained on top. To simulate collecting less data, the first N\%
of contiguous data from each subject was used to train the classifier. This
process introduced an uneven increase in labels per class as the percentage of
training data was increased as reflected by the varying model performance with
respect to the percentage of training data used.

This ECG dataset was in the regime of limited number of subjects and labels
which was prone to overfitting as seen in the fully supervised model trained end
to end; data augmentation and MixUp \cite{zhang_mixup_2018} were applied for
end-to-end training in an attempt to mitigate overfitting. In this scenario,
subject-invariant SSL was important in improving performance. For $\lambda$ of
0.1, the performance for subject-invariant SSL was comparable to the base and
subject-specific SSL up to 40\% of labels and even higher with more labels. By
lowering the regularization ($\lambda$ of 0.01) which increased the amount of
subject-based features in the learned representations, higher accuracies were
achieved.

\begin{figure}
  \centering
  \begin{minipage}[c]{0.68\textwidth}
    \includegraphics[width=\textwidth]{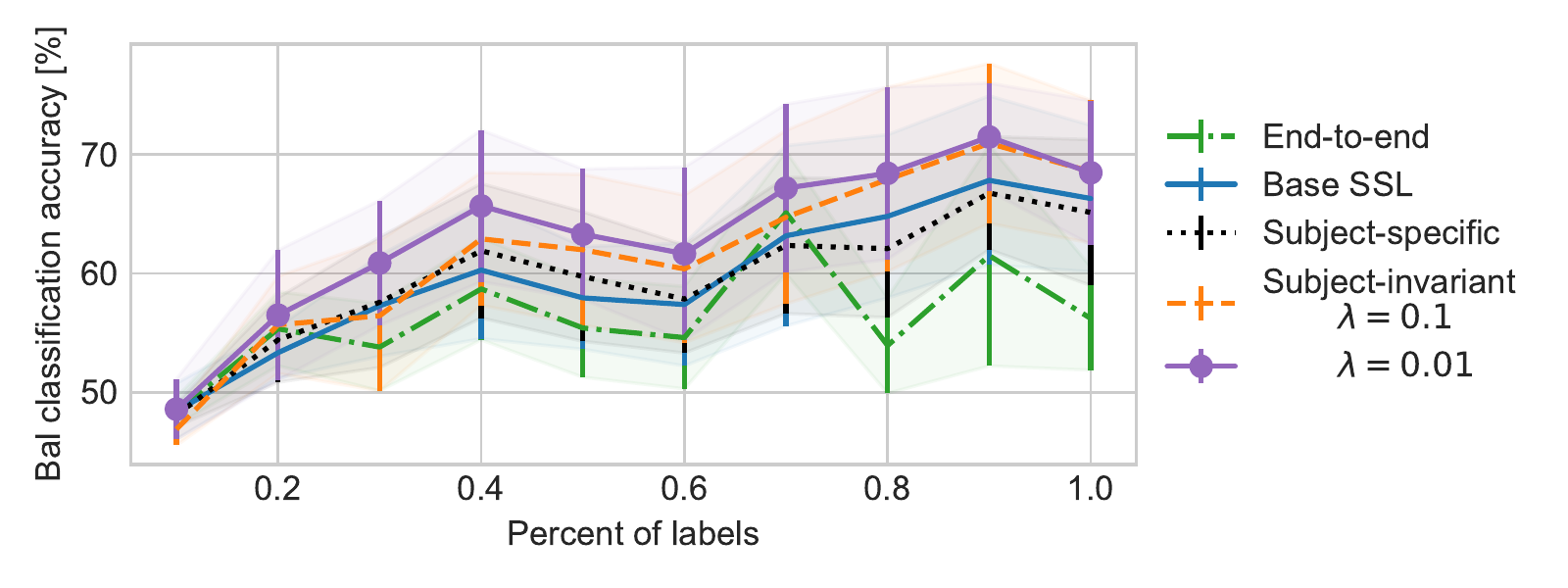}
  \end{minipage}\hfill
  \begin{minipage}[c]{0.32\textwidth}
    \caption{Impact of fewer labels to train a linear classifier on top of the
    frozen encoders for ECG beat classification. Subject information was needed.
    By lowering the degree of subject-invariance ($\lambda=$~0.01), accuracies
    were improved.}
    \label{fig:ecg:labels}
  \end{minipage}
\end{figure}

\paragraph{Fine-Tuning with Supervised Labels.} The self-supervised learned
models were also evaluated as initializations for fine-tuning with supervised
labels. For comparison to previous work \cite{de_chazal_automatic_2004,huang_new_2014,garcia_inter-patient_2017,xu_towards_2019,niu_inter-patient_2019}, weighted
cross-entropy was used \cite{cui_class-balanced_2019} instead of balanced
resampling. We trained an end-to-end supervised model with random initialization
as a baseline for comparison. For beat classification, the baseline model
achieved an overall accuracy of 91.9\%$\pm$1.8\%, F1 score for SVEB of 46.7, and
F1 score for VEB of 89.2. These results were well within the range of previous
work
\cite{de_chazal_automatic_2004,huang_new_2014,garcia_inter-patient_2017,xu_towards_2019,niu_inter-patient_2019}
of 89--96\% overall accuracies, 23--77 F1 for SVEB, and 63--90 F1 for
VEB. Our performance improved by initializing the model with self-supervised
learned weights. The best performance was observed when training from the
subject-specific SSL encoder with an overall accuracy of 93.2\%$\pm$1.6\%, F1
score of SVEB of 43.8, and F1 score of VEB of 92.4. The subject-specific
training produced a model that can be better fine-tuned for the outlier
detection task. See Appendix for more details.


\section{Conclusion}

This work highlights the importance of subject-awareness for learning biosignal
representations. For datasets with a small number of subjects ($<$64 subjects
for EEG), the impact of intersubject variability can be reduced. The
subject-invariant regularization can be reduced for more subjects or if subject
information is important for the downstream task as seen in the analysis with
the ECG dataset.

The work presented can be applied to other biosignals, such as signals from the
eye (EOG) or muscles (EMG), which are influenced by subject-dependent
characteristics. These different data streams are often simultaneously
collected, and self-supervised learning with multimodal data will be an area of
future work. These unlabeled datasets can become many folds larger than the ones
explored, and thus, reducing data requirements and automatically cleaning these
datasets will be important extensions.

Our experiments showed that self-supervised learning, specifically contrastive
learning, provided a solution to handle biosignals. Moreover, minimal
preprocessing was required for these noisy time-series data. With the ease of
collecting unlabeled biosignals, extracting meaningful representations will be
critical in enabling the application of machine learning for personalization and
health.


\section*{Broader Impact}

Electrophysiological sensors are widely used for monitoring, testing, and
diagnosing health conditions. High accuracy and reliability are important when
using machine learning for medical applications. We address the lack of labeled
data and the biases that labeling may introduce to highly noisy time-series
biosignals. However, care must be taken in collecting the unlabeled data to not
bias the learning towards a particular data distribution. The use of
subject-aware training mitigates this concern, but we still recommend
practitioners to check for biases in the learned model. With proper care in data
collection and design, the work presented here enables high quality health
indicators while improving personalization and promoting privacy through
minimization of subject information.


\begin{ack}
  The authors thank Saba Emrani for her advice in working with the MIT-BIH
  Arrhythmia Database of electrocardiogram (ECG) signals. The authors also thank
  Barry Theobald, Russ Webb, Dennis DeCoste, and Nitish Srivastava for helpful
  discussions.
\end{ack}

\bibliographystyle{ieeetr}
\bibliography{jycheng}
\clearpage

\appendix
\counterwithin{figure}{section}
\counterwithin{table}{section}
\section{EEG: PhysioNet Motor Imagery Dataset [Section 4.1]}

\begin{figure}[H]
  \centering
  \includegraphics[scale=0.325]{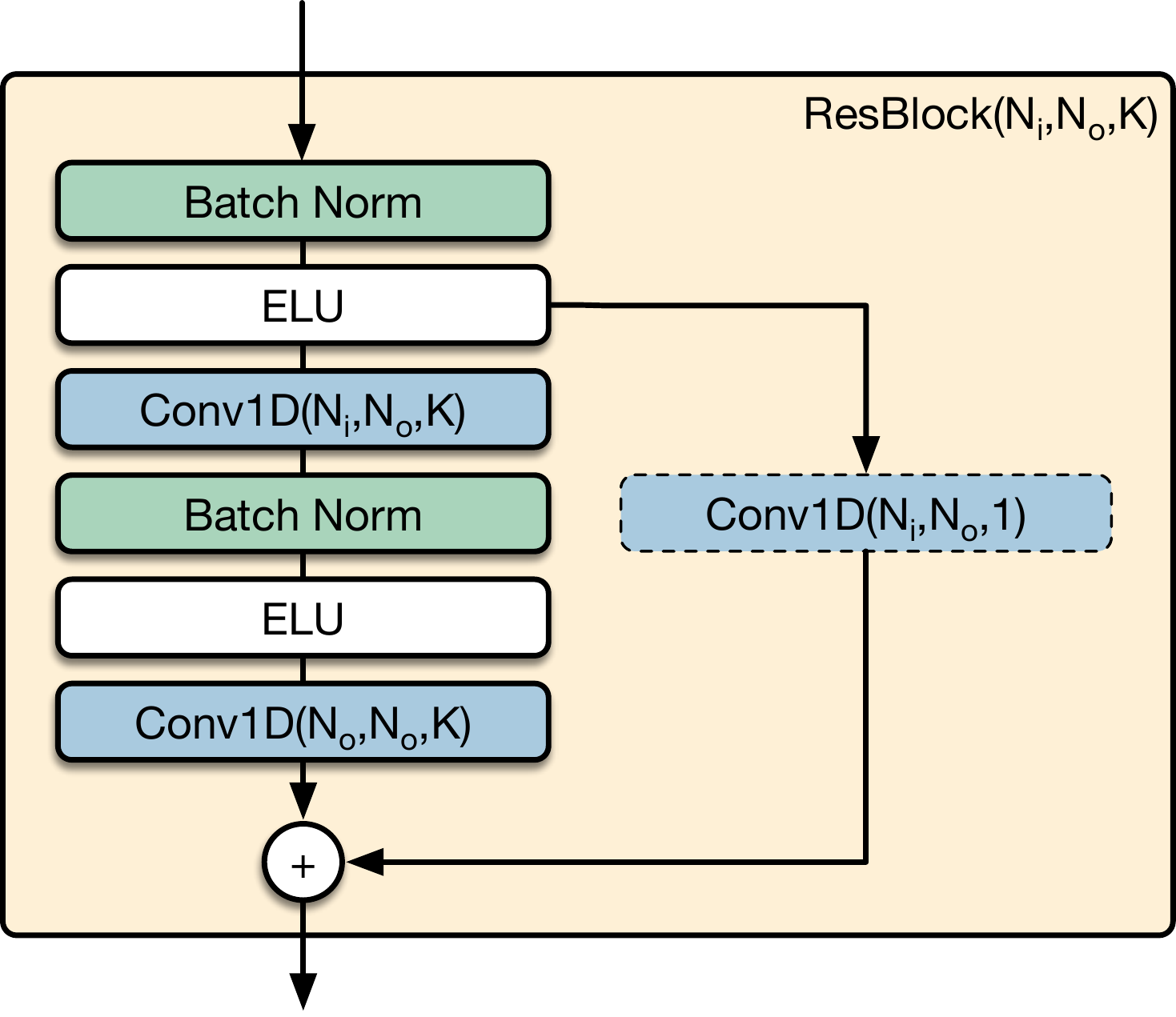}
  \caption{One-dimensional ResNet \cite{he_identity_2016} building block used
  for constructing encoder $G$. This block is built using batch normalization
  \cite{ioffe_batch_2015}, exponential linear unit (ELU) activations, and
  one-dimensional convolutional layers. The convolutional layer is denoted with
  $Conv1D(N_i, N_o, K)$ where $N_i$ and $N_o$ are the number of input and output
  channels, respectively. The kernel size is denoted as $K$. The entire ResBlock
  is denoted as $ResBlock(N_i, N_o, K)$. If $N_i \neq N_o$, a convolutional
  layer (dashed box) is applied in the skip connection.}
  \label{fig:model:resblock}
\end{figure}

\begin{figure}[H]
  \centering
  \includegraphics[scale=0.325]{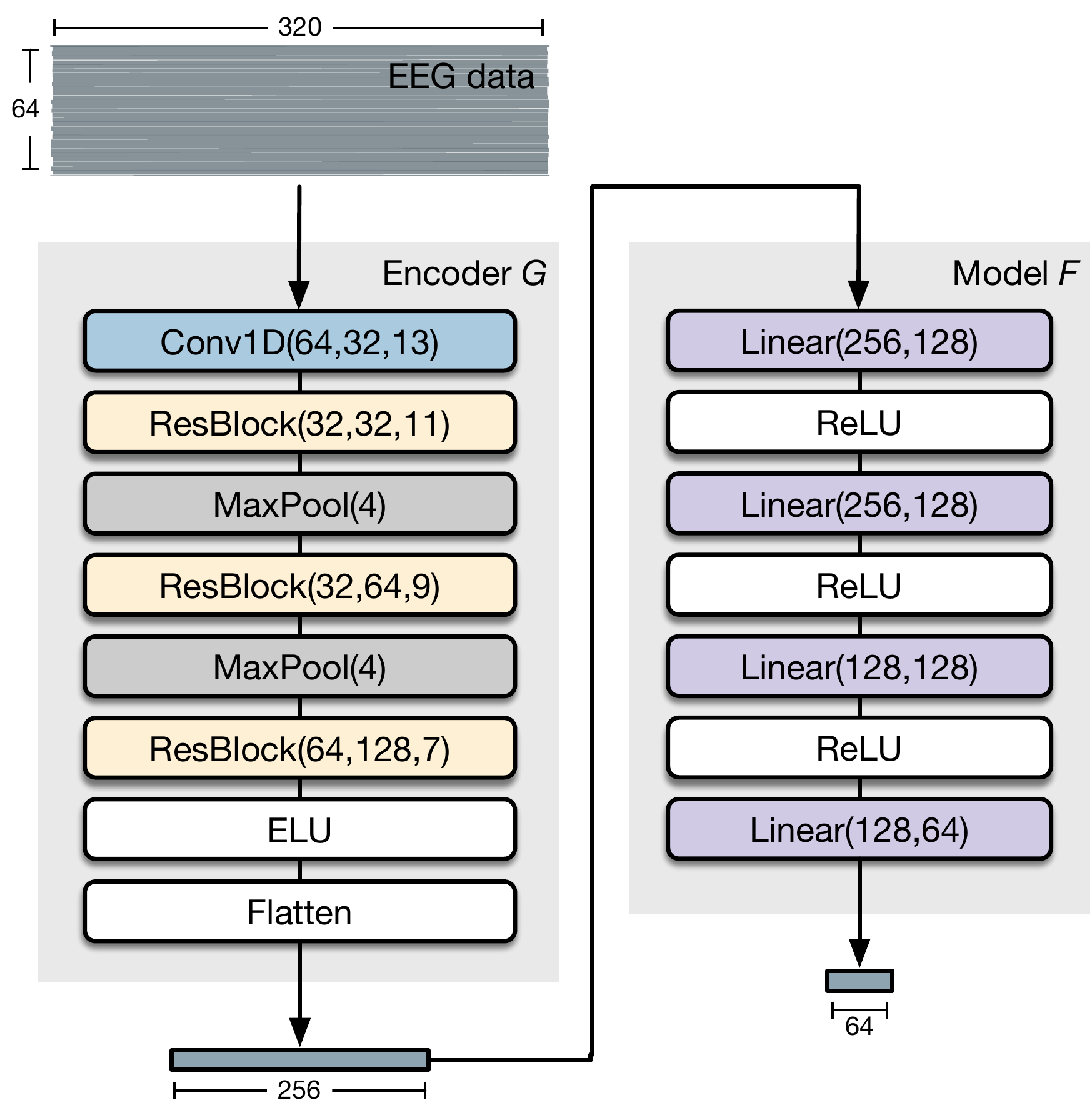}
  \caption{Models used for EEG processing. Model $G$ encodes raw EEG data into a
  256-dimensional vector representation using convolutional layers, ResBlocks
  (Fig.~\ref{fig:model:resblock}), max pooling layers ($MaxPool(K)$ corresponds
  to pooling with a kernel size and stride of $K$), and ELU activation. Model
  $F$ is used during the self-supervised training stage, and this model is built
  using fully connected layers represented with $Linear(N_i, N_o)$ where $N_i$
  and $N_o$ are the input and output dimensions, respectively.}
  \label{fig:model:eeg}
\end{figure}

\paragraph{Model Architecture.} The models used for processing the raw EEG data
are illustrated in Fig.~\ref{fig:model:eeg} using the ResNet building blocks
shown in Fig.~\ref{fig:model:resblock}. The input is 320 samples of the
64-channel data which corresponds to 2.0 sec of data (160 Hz sampling rate).

\begin{figure}[H]
  \centering
  \includegraphics[width=0.49\linewidth]{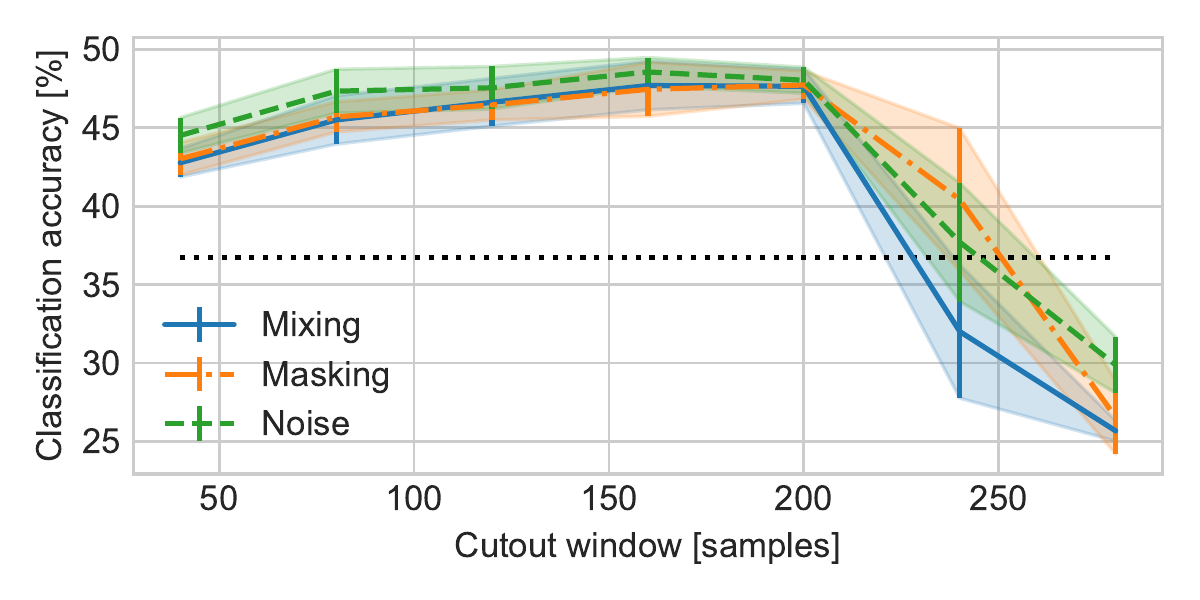}
  \caption{Variations of augmentation demonstrated for cutout performed on the
  EEG PhysioNet Motor Imagery dataset. For random cutouts of the time-series
  signal, the masked region can be replaced with another signal (mixing)
  \cite{yun_cutmix_2019}, zeros (masking) \cite{devries_improved_2017}, or
  Gaussian noise. The width of the cutout window is plotted against the accuracy
  of classifying 4 classes. The data segment is 320 samples wide (2 sec for 160
  Hz sampling rate). Baseline of self-supervised training with no augmentation
  (36.7\%) is plotted in the dotted black line. }
  \label{fig:eeg:aug:cutout}
\end{figure}

\paragraph{Supplementary Results for Impact of Transformation.}
Detailed results for different parameters for each transformation type are
summarized in Table~\ref{table:eeg:augmentations}. Because temporal cutout was
the most effective augmentation task, variants of this transformation were
further investigated. The masked region can be replaced with another signal
(mixing) \cite{yun_cutmix_2019}, zeros \cite{devries_improved_2017}, or Gaussian
noise. From Fig.~\ref{fig:eeg:aug:cutout}, the most effective variant of cutout
was replacing the masked region with noise.  As noted by Chen et
al.~\cite{chen_simple_2020}, stronger augmentations may lead to better learned
representation. We hypothesize that the introduction of noise instead of another
signal increased the difficulty of learning a representation. Though the masked
region was no longer on the data manifold when replaced with noise, this variant
of cutout increased noise in the entire model during training by introducing
more activated neurons throughout the model.


\begin{table}
  \centering
  \caption{Impact of different augmentation parameters on representation
  learning evaluated with a linear classifier trained on top of the learned
  encoder for the PhysioNet EEG Motor Imagery dataset. Accuracies are computed
  for the 4-class problem: imagined motion of left fist, right fist, both fists,
  or feet. These values are computed as mean $\pm$ standard deviation of 10
  trials; the best value is bolded in each row if there is a clear maximum.
  Accuracies are displayed as improvements over baseline of self-supervised
  learning with no augmentation (36.7\%).}
  \label{table:eeg:augmentations}
  \small
  \begin{tabular}{l|llllllllll}
    \toprule
    Max time delay [samples]
    & 40 & 80 & 120 & 160 & 200 & 240 & 280 & 320 \\
    \hspace{5pt}+Accuracy [\%]
    & \textbf{6.1} & 2.4 & -2.2 & -4.6 & -5.9 & -6.2 & -6.7 & -7.3 \\
    & \std{1.2} & \std{1.4} & \std{1.7} & \std{1.6} & \std{1.7} & \std{0.9} & \std{1.6} & \std{0.9} \\
    \midrule

    Time cutout [samples] & 40 & 80 & 120 & 160 & 200 & 240 & 280 \\
    \hspace{5pt}+Accuracy [\%]
    & 5.5 & 8.2 & 9.7 & 10.5 & \textbf{10.9} & 3.1 & -9.5 \\
    & \std{1.3} & \std{0.8} & \std{0.8} & \std{1.8} & \std{2.0} & \std{4.9} & \std{2.3} \\
    \midrule

    Gaussian noise scale
    & 1.0 & 2.0 & 3.0 & 4.0 & 5.0 & 6.0 & 7.0 & 8.0 & 9.0 & 10.0 \\
    \hspace{5pt}+Accuracy [\%]
    & 1.3 & 2.2 & 2.3 & 2.9 & 3.5 & \textbf{3.8} & 3.3 & 3.1 & 3.2 & 3.7 \\
    & \std{1.2} & \std{1.7} & \std{1.2} & \std{1.2} & \std{0.8} & \std{0.9} & \std{1.5} & \std{1.3} & \std{1.5} & \std{1.4} \\
    \midrule

    Bandstop width [Hz]
    & 1 & 2 & 4 & 8 & 16 & 32 & 64 & 96 \\
    \hspace{5pt}+Accuracy [\%]
    & 1.5 & 1.6 & 1.7 & 2.1 & 2.0 & 3.0 & \textbf{3.7} & 2.5 \\
    & \std{1.4} & \std{1.3} & \std{1.4} & \std{1.0} & \std{2.9} & \std{2.0} & \std{1.5} & \std{1.5} \\

    \midrule
    Signal mixing scale
    & 0.1 & 0.2 & 0.3 & 0.4 & 0.5 & 0.6 & 0.7 & 0.8 & 0.9 & 1.0 \\
    \hspace{5pt}+Accuracy [\%]
    & 2.4 & 2.1 & 2.5 & 2.6 & 2.7 & 2.6 & 3.1 & 4.1 & \textbf{5.3} & -11.0 \\
    & \std{1.0} & \std{0.8} & \std{1.0} & \std{1.0} & \std{0.7} & \std{0.7} & \std{1.3} & \std{1.2} & \std{1.4} & \std{0.6} \\
    \midrule

    Max spatial shift$^a$
    & 0.02 & 0.04 & 0.06 & 0.08 & 0.1 & 0.12 & 0.14 & 0.16 & 0.18 \\
    \hspace{5pt}+Accuracy [\%]
    & \textbf{0.1} & -0.3 & -0.3 & -0.2 & -0.1 & -0.3 & -0.5 & -0.0 & -0.3 \\
    & \std{1.1} & \std{0.6} & \std{1.4} & \std{1.1} & \std{1.0} & \std{0.7} & \std{1.0} & \std{1.1} & \std{1.3} \\
    \midrule

    Max rotation [$^\circ$]
    & 10 & 20 & 30 & 40 & 50 & 60 & 70 & 80 & 90 \\
    \hspace{5pt}+Accuracy [\%]
    & -0.5 & -0.1 & 0.0 & -0.3 & -1.2 & -0.6 & -0.2 & -0.1 & -1.0 \\
    & \std{1.2} & \std{0.9} & \std{0.7} & \std{1.5} & \std{1.1} & \std{1.7} & \std{0.0} & \std{0.7} & \std{1.5} \\
    \midrule

    Sensor dropout $p$
    & 0.1 & 0.2 & 0.3 & 0.4 & 0.5 & 0.6 & 0.7 & 0.8 & 0.9 \\
    \hspace{5pt}+Accuracy [\%]
    & 2.8 & \textbf{3.8} & 3.6 & 3.2 & 2.2 & 0.3 & -0.5 & -3.2 & -3.4 \\
    & \std{1.3} & \std{1.5} & \std{1.4} & \std{0.9} & \std{1.0} & \std{1.5} & \std{1.2} & \std{1.4} & \std{1.6} \\
    \midrule

    Sensor cutout radius$^a$
    & 0.05 & 0.10 & 0.15 & 0.20 & 0.25 & 0.30 & 0.35 & 0.40 & 0.45 \\
    \hspace{5pt}+Accuracy [\%]
    & 1.6 & 1.5 & 3.3 & 3.9 & \textbf{4.1} & 2.8 & 1.6 & 0.1 & -0.7 \\
    & \std{1.1} & \std{1.4} & \std{1.0} & \std{0.9} & \std{1.4} & \std{1.3} & \std{1.2} & \std{1.5} & \std{1.5} \\
    \bottomrule
  \end{tabular}\\
  \footnotesize{$^a$Spatial units are normalized such that the entire extent of the coverage in all directions is between 0 and 1.}
\end{table}

\begin{figure}[H]
  \centering
  \begin{subfigure}[t]{0.49\textwidth}
    \includegraphics[width=\linewidth]{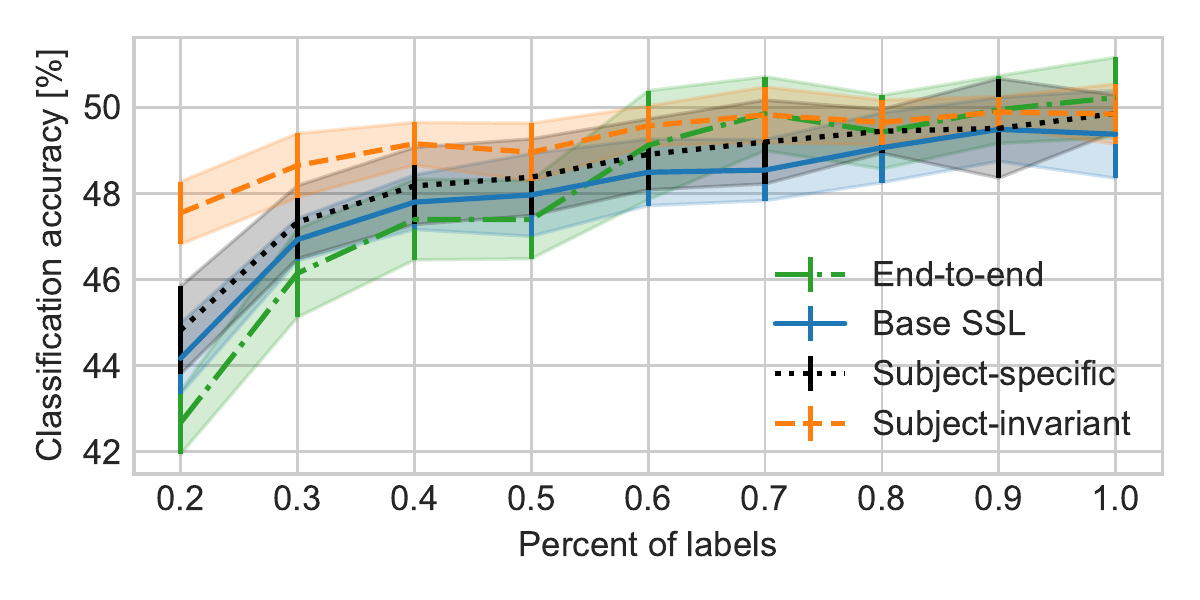}
    \caption{\textit{Intersubject testing:} train on 90 subjects (also used for
    self-supervised learning) and test on 16 unseen subjects. The first $N$\% of
    runs ($x$-axis) for each subject were used for training the linear
    classifier.}
  \end{subfigure}
  \hfill
  \begin{subfigure}[t]{0.49\textwidth}
    \includegraphics[width=\linewidth]{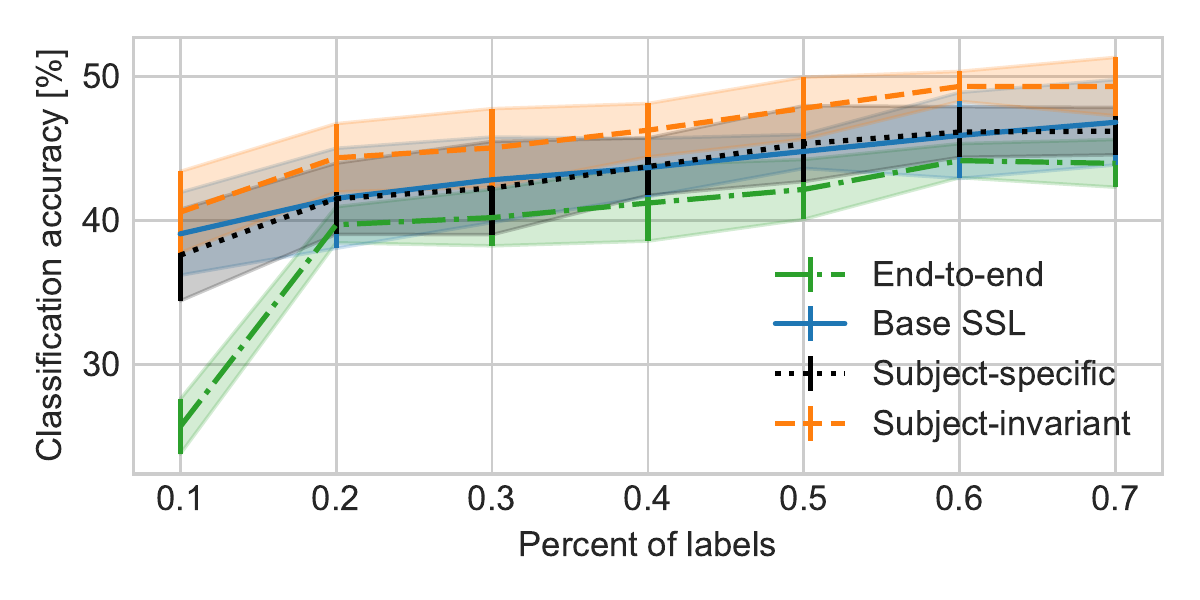}
    \caption{\textit{Intrasubject testing:} training and testing on 16 subjects
    that were not used for self-supervised training. The first $N$\% of runs
    ($x$-axis) for each subject were used for training and the last 25\% for testing.}
  \end{subfigure}
  \caption{Classification accuracy for 4 classes on the EEG dataset. A linear
  classifier was trained on top of the frozen encoder. Supervised end-to-end
  learning was used as a baseline for reference. With limited labels ($<$0.7 for
  intersubject and any percentage for intrasubject), the subject-invariant
  consistently out-performed all other approaches due to its ability to limit
  intersubject variability.}
  \label{fig:eeg:labels:supp}
\end{figure}

\paragraph{Training Details for Fine-Tuning with Supervised Labels.}
The models produced from self-supervised learning were evaluated as an
initialization for supervised fine-tuning. More specifically, a
randomly-initialized fully connected layer was attached to the end of the
encoder, and the fully connected layer and the encoder were updated during the
training. Adam optimizer with a learning rate of 1e-5 was used with a batch size
of 256 for 200 epochs. For comparison, we also trained a network end-to-end from
random initialization with a learning rate of 1e-2, batch size of 1000, and 1k
epochs. In all cases, data augmentation and MixUp \cite{zhang_mixup_2018} were
used to mitigate overfitting. A representative confusion matrix for the
subject-invariant model is shown in Table~\ref{table:eeg:confusion}.

\begin{table}[H]
  \caption{Representative confusion matrix for 4-way classification for EEG
  decoding using model trained with subject-invariant learning. This model was
  fine-tuned with supervised labels. The overall accuracy was 54.4\%.}
  \label{table:eeg:confusion}
  \centering
  \begin{tabular}{rr|rrrr}
    \toprule
    & & \multicolumn{4}{c}{Predicted Label} \\
    & & Left & Right & Fists & Feet \\
    \hline
    \parbox[t]{2mm}{\multirow{4}{*}{\rotatebox[origin=c]{90}{True label}}}
    &  Left & 230 &  28 &  67 &  38 \\
    & Right &  36 & 208 &  48 &  63 \\
    & Fists &  69 &  54 & 169 &  66 \\
    &  Feet &  46 &  54 &  87 & 175 \\
    \bottomrule
  \end{tabular}
\end{table}
\clearpage
\section{ECG: MIT-BIH Arrhythmia Database [Section 4.2]}

\begin{figure}[H]
  \centering
  \includegraphics[scale=0.325]{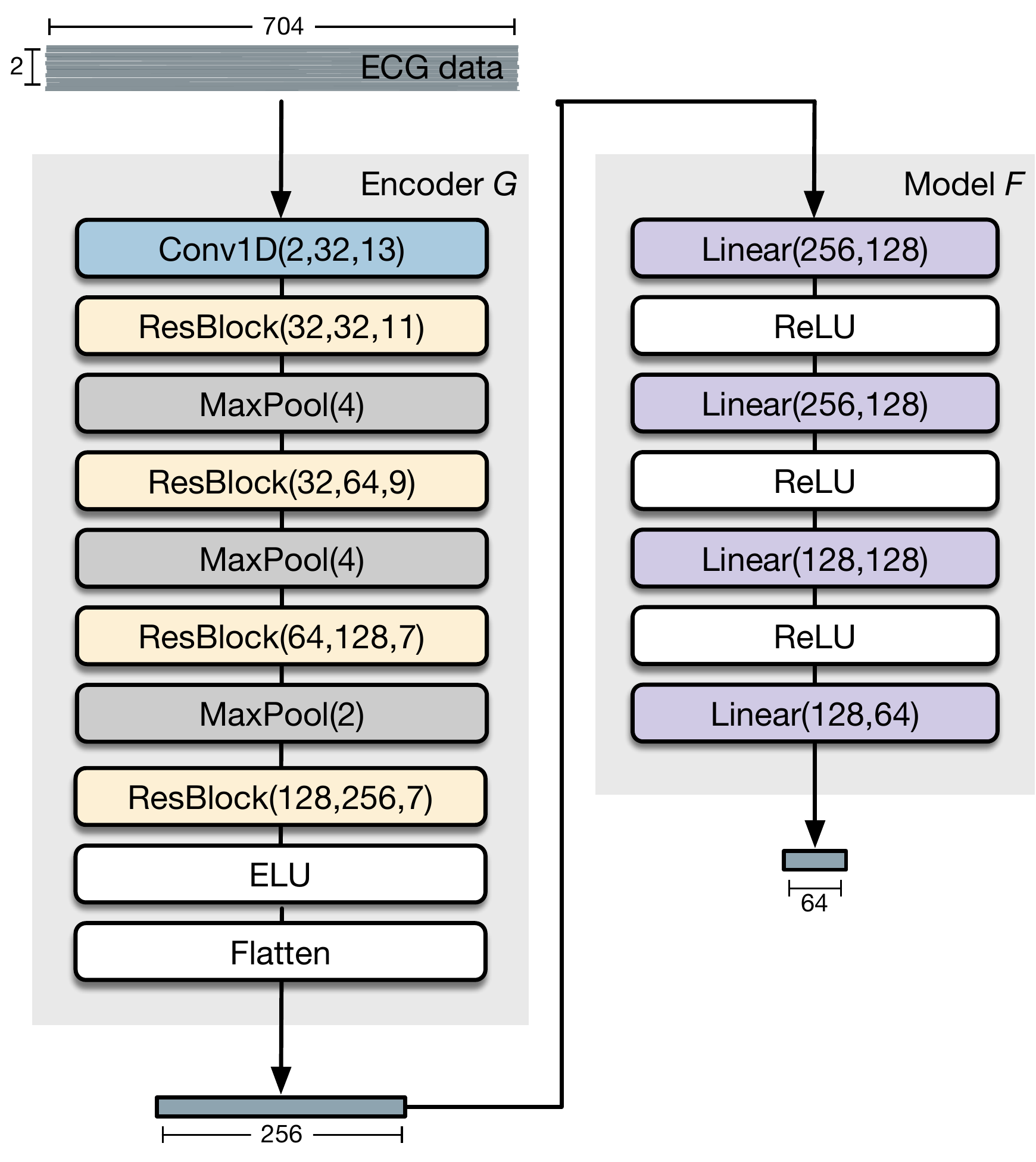}
  \caption{Models used for ECG processing. Model $G$ encodes raw ECG data into a
  256-dimensional vector representation. Model $F$ is used during the
  self-supervised training stage.}
  \label{fig:model:ecg}
\end{figure}

\paragraph{Model Architecture.} The models used for processing the raw ECG data
are illustrated in Fig.~\ref{fig:model:ecg} using the ResNet building blocks
shown in Fig.~\ref{fig:model:resblock}. The input is 704 samples of the
2-channel data which corresponds to 1.96 sec of data (360 Hz sampling
rate).

\paragraph{Training Details for Fine-Tuning with Supervised Labels.}
The self-supervised learned models were also evaluated as an initialization for
supervised training. In literature
\cite{de_chazal_automatic_2004,huang_new_2014,garcia_inter-patient_2017,xu_towards_2019,niu_inter-patient_2019},
the accuracies of classifying SVEB and VEB were given higher importance. For
comparison, weighted cross-entropy was used \cite{niu_inter-patient_2019}
instead of balanced resampling. Weights were computed using the following
formulation \cite{cui_class-balanced_2019}: $w[i] = (1-\beta^{n[i]})/(1-\beta)$
where $n[i]$ is the number of samples for class $i$, and $\beta$ is a
hyperparameter set to 0.999. Due to limited number of samples for the unknown
beat class ($<$20 in training and test set combined), this class was essentially
ignored by assigning the class weight to be the same as the majority class.

One important feature for beat classification is the current RR interval (period
of a single heart beat) along with the RR intervals of neighboring heart beats
\cite{doquire_feature_2011,lin_heartbeat_2014}. This feature is unavailable if
the receptive field does not cover multiple heart beats as in our setup. The
learned representation can be enhanced by including this domain-specific
feature. Following previous work \cite{huang_new_2014,niu_inter-patient_2019},
the RR intervals before and after the current heart beat were calculated and
normalized by the average RR interval over each recording. This RR information
was concatenated to the embedding vector produced by the encoder network.

Two fully-connected layers (with 128 dimensions in the hidden layer) were
attached to the output of encoder $G$. The encoder $G$ was initialized with the
model produced from self-supervised learning. The entire model was then trained
end to end during the supervised training. Adam optimizer with a learning rate
of 1e-5 was used to update the model with a batch size of 256 for 50 epochs. For
comparison, a randomly initialized model was trained end to end with a learning
rate of 1e-3, batch size of 256, and 100 epochs. Results are summarized in
Table~\ref{table:ecg:beat:fine}.

The inclusion of the RR interval improved the end-to-end supervised
classification performance from F1 score for SVEB of 31.4 to 46.7 and for VEB of
85.8 to 89.2 --- well within the range of previous work
\cite{de_chazal_automatic_2004,huang_new_2014,garcia_inter-patient_2017,xu_towards_2019,niu_inter-patient_2019}
of 23.2--76.6 for SVEB and 63.4--89.7 for VEB. For beat classification from ECG,
we observed the best performance when training from the subject-specific SSL
encoder with an overall accuracy of 93.2\%$\pm$1.6\%, F1 score of SVEB of 43.8,
and F1 score of VEB of 92.4. A representative confusion matrix for
subject-specific SSL is shown in Table~\ref{table:ecg:confusion}.


\begin{table}[H]
  \caption{Accuracies for intersubject beat classification: the training and
  testing were performed on separate set of subjects. General representations
  were learned through self-supervised training, and fine-tuning from these
  models improved accuracies (last six rows). For conciseness, only the standard
  deviation of the overall accuracies and F1 scores are reported. Our best
  results are bolded.}
  \label{table:ecg:beat:fine}
  \centering
  \begin{tabular}{lccccc}
    \toprule
    & \multicolumn{1}{c}{Overall}
    & \multicolumn{2}{c}{SVEB}
    & \multicolumn{2}{c}{VEB} \\
    \cmidrule(lr){2-2}\cmidrule(lr){3-4}\cmidrule(lr){5-6}
    & Acc & Acc / Se / +P & F1 & Acc / Se / +P & F1 \\
    \hline
    Luo et al.~\cite{luo_patient-specific_2017} & 89.3
    & 96.2 / 15.4 / 47.3 & 23.2 & 95.5 / 60.4 / 66.8 & 63.4 \\
    Huang et al.~\cite{huang_new_2014} & 93.8
    & 95.1 / 91.1 / 42.2 & 57.7 & 99.0 / 93.9 / 90.9 & 92.4\\
    Niu et al.~\cite{niu_inter-patient_2019} & 96.4
    & - / 76.5 / 76.6 & 76.6 & - / 85.7 / 94.1 & 89.7 \\
    \hline
    End-to-end (ours)
    & 91.0\std{2.0}
    & 95.9 / 25.8 / 41.8 & 31.4\std{10.3}
    & 97.8 / 95.6 / 78.9 & 85.8\std{8.2}\\

    End-to-end$^{R}$ (ours)
    & 91.9\std{1.8}
    & 96.3 / 44.5 / 52.2 & \textbf{46.7\std{14.9}}
    & 98.4 / 96.4 / 83.4 & 89.2\std{4.3} \\

    \hline
    \textit{Fine-tuned from:}\\
    \hspace{10pt}Base SSL$^{R}$
    & 91.7\std{1.8}
    & 95.6 / 30.2 / 39.2 & 33.3\std{8.1}
    & 98.5 / 96.8 / 83.7 & 89.6\std{3.9} \\

    \hspace{10pt}Subject-specific$^{R}$
    & \textbf{93.2\std{1.6}}
    & 96.0 / 42.8 / 50.1 & 43.8\std{10.0}
    & 98.9 / 96.6 / 88.7 & \textbf{92.4\std{3.3}} \\

    \hspace{10pt}Subject-invariant$^{R}$\\
    \hspace{15pt}$\lambda=$~1.0
    & 89.4\std{3.0}
    & 96.3 / 38.8 / 48.9 & 42.5\std{13.8}
    & 97.8 / 95.8 / 77.4 & 85.4\std{4.4} \\

    \hspace{15pt}$\lambda=$~0.1
    & 91.8\std{2.2}
    & 96.2 / 35.5 / 48.8 & 40.8\std{14.6}
    & 98.0 / 95.3 / 79.2 & 86.3\std{5.3} \\

    \hspace{15pt}$\lambda=$~0.01
    & 90.4\std{2.2}
    & 95.7 / 35.2 / 42.2 & 37.3\std{16.8}
    & 98.1 / 96.4 / 80.0 & 87.2\std{5.2} \\


    \hspace{15pt}$\lambda=$~0.001
    & 91.3\std{2.7}
    & 95.8 / 26.5 / 40.1 & 31.6\std{13.5}
    & 98.1 / 96.2 / 80.4 & 87.2\std{7.1} \\

    \bottomrule
  \end{tabular}\\
  \footnotesize{$^R$Including RR-interval information as input to the
  classifier.\\
  Abbreviations: SVEB~=~supraventricular ectopic beat; VEB~=~ventricular ectopic
  beat; Acc~=~accuracy [\%]; Se~=~sensitivity [\%]; +P~=~positive
  predictability; F1~=~F1 score.}
\end{table}

\begin{table}[H]
  \caption{Representative confusion matrix for beat classification using model
  trained with subject-specific contrastive learning. This model was fine-tuned
  with supervised labels. The overall accuracy was 95.1\%. F1 scores for SVEB
  and VEB were 46.1 and 94.6, respectively.}
  \label{table:ecg:confusion}
  \centering
  \begin{tabular}{rr|rrrrr}
    \toprule
    & & \multicolumn{5}{c}{Predicted Label} \\
    & &     N & SVEB &  VEB &   F & Q \\
    \hline
    \parbox[t]{2mm}{\multirow{5}{*}{\rotatebox[origin=c]{90}{True label}}}
    &    N & 43376 &  516 &  115 & 193 & 1 \\
    & SVEB &  1056 &  717 &   59 &   5 & 0 \\
    &  VEB &    60 &   39 & 3067 &  53 & 0 \\
    &    F &   299 &    0 &   18 &  71 & 0 \\
    &    Q &     3 &    0 &    4 &   2 & 0 \\
    \bottomrule
  \end{tabular}\\
  \footnotesize{Abbreviations: N~=~normal beat; SVEB~=~supraventricular ectopic beat; VEB~=~ventricular ectopic beat; F~=~fusion beat; Q~=~unknown beat.}
\end{table}

\end{document}